\documentclass[conference]{IEEEtran}
\IEEEoverridecommandlockouts

\usepackage[colorlinks]{hyperref}
\usepackage{amsmath,graphicx,booktabs,multirow,colortbl}
\usepackage{xcolor}
\usepackage{bbm}
\usepackage{amssymb} 
\usepackage{pifont}  
\newcommand{\Checkmark}{\checkmark}    

\def\BibTeX{{\rm B\kern-.05em{\sc i\kern-.025em b}\kern-.08em
    T\kern-.1667em\lower.7ex\hbox{E}\kern-.125emX}}
\begin{document}

\title{DFR: A Decompose-Fuse-Reconstruct Framework for Multi-Modal Few-Shot Segmentation}
\author{\IEEEauthorblockN{Shuai Chen, Fanman Meng\textsuperscript{*}, Xiwei Zhang, Haoran Wei, Chenhao Wu, Qingbo Wu, Hongliang Li}
\IEEEauthorblockA{\textit{School of Information and Communication Engineering} \\
\textit{University of Electronic Science and Technology of China}\\
Chengdu, China\
\{schen, 202322011832, hrwei, chwu\}@std.uestc.edu.cn, \{fmmeng, qbwu, hlli\}@uestc.edu.cn}
\thanks{\textsuperscript{*}Corresponding Author}}

\maketitle

\begin{abstract}

This paper presents DFR (Decompose, Fuse and Reconstruct), a novel framework that addresses the fundamental challenge of effectively utilizing multi-modal guidance in few-shot segmentation (FSS). While existing approaches primarily rely on visual support samples or textual descriptions, their single or dual-modal paradigms limit exploitation of rich perceptual information available in real-world scenarios. To overcome this limitation, the proposed approach leverages the Segment Anything Model (SAM) to systematically integrate visual, textual, and audio modalities for enhanced semantic understanding. The DFR framework introduces three key innovations: 1) Multi-modal Decompose: a hierarchical decomposition scheme that extracts visual region proposals via SAM, expands textual semantics into fine-grained descriptors, and processes audio features for contextual enrichment; 2) Multi-modal Contrastive Fuse: a fusion strategy employing contrastive learning to maintain consistency across visual, textual, and audio modalities while enabling dynamic semantic interactions between foreground and background features; 3) Dual-path Reconstruct: an adaptive integration mechanism combining semantic guidance from tri-modal fused tokens with geometric cues from multi-modal location priors. Extensive experiments across visual, textual, and audio modalities under both synthetic and real settings demonstrate DFR's substantial performance improvements over state-of-the-art methods.

\end{abstract}


\section{Introduction}
\label{sec:intro}
Semantic segmentation serves as a cornerstone for visual scene understanding, with deep learning approaches~\cite{long2015fully,chen2017deeplab,shi2022ssformer} achieving remarkable success through large-scale supervised training. Despite these advances, the requirement for extensive pixel-wise annotations poses significant challenges when generalizing to novel categories. Therefore, Few-shot segmentation (FSS) emerges as a promising paradigm to address this limitation by learning to segment unseen categories from limited labeled examples.

\begin{figure}
  \centering
  \includegraphics[width=0.99\linewidth]{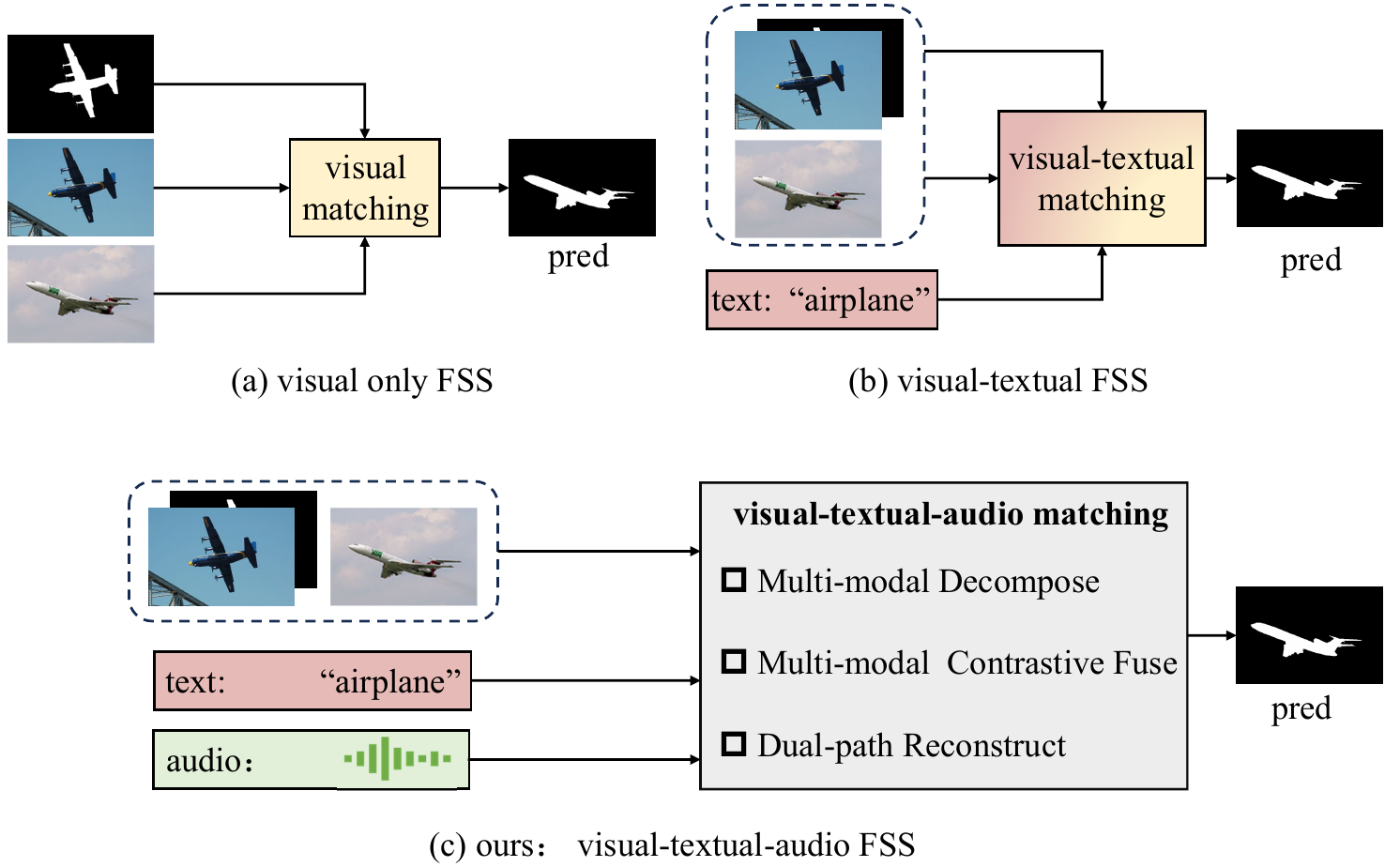}
  \caption{Illustration of evolution of FSS frameworks: from visual-only/visual-textual paradigms to our proposed multi-modal decomposition-fusion-reconstruction architecture incorporating audio signals.}
  \label{fig:motivation}
\end{figure}

Recent progress in FSS has witnessed an evolution from purely visual approaches~\cite{tian2020prior, min2021hypercorrelation,herzog2024adapt,Wang_2023_CVPR,bao2024relevant,li2024label} to visual-textual based frameworks~\cite{luddecke2022image}, demonstrating the effectiveness of leveraging linguistic semantics~\cite{radford2021learning} for generalization. As illustrated in Figure~\ref{fig:motivation}, while existing methods have predominantly focused on either visual-only or visual-textual paradigms, real-world scenarios inherently contain rich perceptual information beyond these modalities. Particularly, audio signals~\cite{10647485}, which encode temporal-dynamic characteristics and object-specific acoustic patterns, remain largely unexplored in FSS despite their potential to provide complementary semantic cues. This observation motivates us to develop a comprehensive multi-modal few-shot segmentation (MMFSS) framework that systematically integrates audio information with visual and textual modalities, as depicted in the bottom part of Figure~\ref{fig:motivation}.

The integration of multiple heterogeneous modalities for FSS presents two fundamental challenges. First, different modalities exhibit distinct structural characteristics, i.e., visual features are spatially organized and fine-grained, textual embeddings capture hierarchical semantics (category, attributes, and context), and audio signals encode temporal-frequency patterns. Establishing effective correspondence across these heterogeneous representations while preserving modality-specific discriminative properties requires careful architectural design. Second, conventional multi-modal fusion strategies face unique challenges in few-shot scenarios, where maintaining semantic consistency across modalities becomes particularly crucial yet difficult due to limited training samples. This limitation necessitates a principled approach to align and validate cross-modal feature representations while maximizing the utility of sparse labeled data.

We address these challenges through DFR, built upon the foundation of SAM's~\cite{kirillov2023segment} powerful visual understanding and LanguageBind's~\cite{lbz2024} cross-modal alignment capabilities. Our approach introduces three key innovations: (1) a multi-modal decomposition scheme that systematically extracts and enriches features across modalities through SAM-based region proposals, LLM-guided semantic expansion, and AudioLDM-generated acoustic embeddings; (2) a contrastive fusion mechanism that maintains modality consistency through InfoNCE loss while enabling dynamic interactions between foreground and background features; and (3) a dual-path reconstruction module that adaptively integrates semantic tokens with geometric prompts derived from multi-modal location priors. Our primary contributions are:

\begin{itemize}
  \item A novel multi-modal FSS framework that systematically integrates and aligns visual, textual, and audio modalities through a unified architecture, establishing a new paradigm for real-world segmentation tasks. 
  \item A hierarchical decomposition and progressive fusion mechanism that enables fine-grained cross-modal feature learning while preserving modality-specific characteristics through contrastive regularization.
  \item Extensive validation demonstrates DFR's substantial performance gains across both synthetic and real audio settings, achieving 7.3\% and 2.2\% mIoU improvements (1-shot and 5-shot) on PASCAL-5i with synthetic audio, and 4.8\% and 3.3\% mIoU improvements (0-shot and 1-shot) on real audio-visual segmentation dataset AVS-V3.

\end{itemize}

\section{Related Work}
\label{sec:related_work}

\subsection{Few-Shot Segmentation}
Few-shot segmentation approaches can be categorized into three main paradigms based on their guidance modalities: visual-only methods, visual-textual methods, and multi-modal methods. Visual-guided methods, serving as the default paradigm in FSS, typically follow either prototype-based or matching-based frameworks. Prototype-based methods~\cite{wang2019panet,tian2020prior} focus on extracting class-specific representations from support images, evolving from simple global prototypes to more sophisticated multiple prototype systems. Matching-based approaches~\cite{min2021hypercorrelation} establish dense pixel-level correspondences between support and query features, enabling better preservation of spatial details. 

Recent advances have introduced textual modality as complementary guidance, marking a significant shift towards multi-modal understanding. Methods like~\cite{luddecke2022image} leverage vision-language models~\cite{radford2021learning} to enhance generalization to novel categories through semantic alignment. While these visual-textual methods demonstrate improved performance over visual-only approaches, they are inherently limited to bi-modal interactions. The potential of other modalities, particularly audio signals which encode object-specific temporal-dynamic patterns, remains largely unexplored in FSS. This observation aligns with our motivation to develop a more comprehensive multi-modal framework that leverages the complementary strengths of visual, textual, and audio modalities.

\subsection{Segment Anything Model in FSS}
The Segment Anything Model (SAM)~\cite{kirillov2023segment} has emerged as a powerful foundation for segmentation tasks through its prompt-based architecture and zero-shot generalization capabilities. Its ability to decompose images into meaningful region proposals naturally aligns with FSS requirements. Recent works have explored various strategies to leverage this synergy: VRP-SAM~\cite{sun2024vrp} introduces a visual reference prompt encoder to automatically generate prompts from reference images, Matcher~\cite{liu2023matcher} achieves impressive results through training-free bidirectional matching and robust prompt sampling, while FCP~\cite{fcp} develops a foreground-covering prototype generation approach. However, these methods primarily focus on visual prompt engineering, leaving the potential of multi-modal prompts largely unexplored. Our work bridges this gap by introducing a dual-path reconstruction mechanism that combines SAM's geometric understanding with rich semantic cues from multiple modalities.

\begin{figure*}[t]
  \centering
  \includegraphics[width=1.0\linewidth]{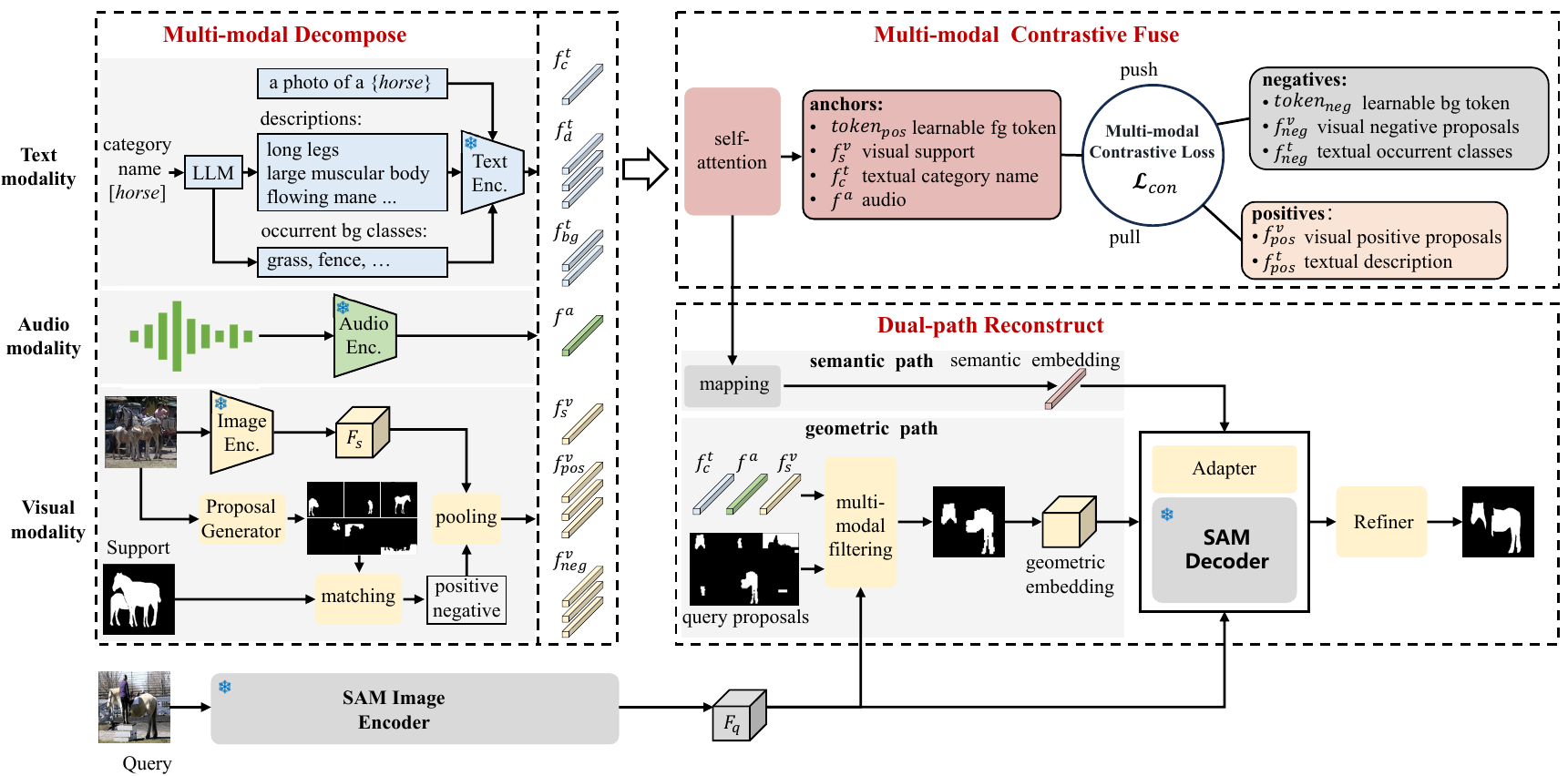}
  \caption{Overview of the proposed Decompose-Fuse-Reconstruct (DFR) framework for multi-modal few-shot segmentation. Our approach consists of three key stages: (i) Multi-modal Decompose: hierarchically extracting features through SAM-based region proposals, LLM-guided semantic expansion, and audio embeddings, (ii) Multi-modal Contrastive Fuse: maintaining modality consistency while enabling dynamic foreground-background interactions through InfoNCE-based regularization, and (iii) Dual-path Reconstruct: adaptively integrating semantic guidance with geometric cues from multi-modal location priors for precise segmentation.}
  \label{fig:framework}
\end{figure*}

\section{Proposed method}
\label{sec:method}

\subsection{Problem Formulation}

Few-shot segmentation tackles the fundamental challenge of generalizing segmentation capabilities to novel categories with minimal supervision. Let $\mathcal{C}_{base}$ and $\mathcal{C}_{novel}$ denote the base and novel categories respectively, where $\mathcal{C}_{base} \cap \mathcal{C}_{novel} = \emptyset$. During training, the model has access to abundant labeled samples from base categories, while during testing, it needs to segment objects from novel categories with only a few support examples. Formally, in the K-shot setting, each episode consists of a support set $\mathcal{S} = \{(I_s^i, M_s^i)\}_{i=1}^K$ containing K image-mask pairs and a query image $I_q \in \mathbb{R}^{H \times W \times 3}$ to be segmented, where traditional FSS methods aim to learn a mapping function $M_q = \Phi(I_q, \mathcal{S})$. In this work, we extend the conventional FSS formulation into multi-modal few-shot segmentation (MMFSS) by incorporating multi-modal guidance. Specifically, for each category we introduce additional textual category name $T$ and audio signals $A$ that provide complementary semantic cues, formulating an enhanced few-shot segmentation task as $M_q = \Phi(I_q, \mathcal{S}, T, A)$.

\subsection{DFR Framework}

Figure~\ref{fig:framework} presents our proposed DFR framework, which systematically integrates multi-modal information through three key stages: decomposition, fusion, and reconstruction. The framework is built upon the foundation of SAM while incorporating novel modules for multi-modal processing.

\subsubsection{Multi-modal Decompose}
Few-shot segmentation requires rich semantic understanding across modalities. However, conventional approaches often suffer from information loss due to oversimplified representations. To address this, we propose a Multi-modal decomposition scheme that systematically disentangles and enriches representations across visual, textual, and audio modalities.

\noindent\textbf{Visual Decomposition.} 
We leverage SAM to decompose support images into region proposals $\mathcal{P} = \{P_i\}_{i=1}^N$. These proposals are categorized based on their overlap with support mask $M_s$ using overlap ratio:
\begin{equation}
  \text{OR}(P_i, M_s) = \frac{|P_i \cap M_s|}{|P_i|},
\end{equation}
wher proposals with $\text{OR} > \tau$ ($\tau=0.5$) form positive set $\mathcal{P}^{+}$, others form negative set $\mathcal{P}^{-}$. This enables derivation of three visual prototypes via pooling on extracted support features $F_s$: positive prototype $f_{pos}^v$, negative prototype $f_{neg}^v$, and support prototype $f_s^v$.

\noindent\textbf{Textual Decomposition.} 
To enrich semantic understanding beyond category labels, we employ large language models to generate comprehensive textual representations. For each category, we extract three types of semantic features: (1) category name embedding $f_c^t$, (2) fine-grained descriptive attributes embedding $f_d^t$ obtained through prompting: \textit{"For an image containing [category], what features distinguish it from other potentially co-existing categories?"}, and (3) background context embedding $f_{bg}^t$ derived from LLM's answer of potentially co-existing categories in the scene.

\noindent\textbf{Audio Decomposition.} 
We utilize AudioLDM~\cite{audioLDM} to synthesize characteristic sound effects $\mathcal{A} = \text{AudioLDM}(\mathcal{T})$, which are processed to obtain audio embedding $f^a$, providing complementary temporal-dynamic information.

\subsubsection{Multi-modal Contrastive Fuse}
Multi-modal fusion faces two challenges: integrating heterogeneous features while maintaining modality-specific characteristics, and distinguishing target semantics from background interference. We propose a contrastive fusion strategy to address these challenges.

Our fusion process combines foreground features $f_{fg} = [token_{pos}; f_s^v; f_{pos}^v; f_c^t; f_d^t; f^a]$ and background features $f_{bg} = [token_{neg}; f_{neg}^v; f_{bg}^t]$, where $token_{pos}$ and $token_{neg}$ are learnable tokens. These features are enhanced through self-attention:
\begin{equation}
  \begin{aligned}
  f_{pos} &= \text{softmax}\left(\frac{f_{pos}\mathbf{W}_Q^p(f_{pos}\mathbf{W}_K^p)^\top}{\sqrt{d_k}}\right)f_{pos}\mathbf{W}_V^p, \\
  f_{neg} &= \text{softmax}\left(\frac{f_{neg}\mathbf{W}_Q^n(f_{neg}\mathbf{W}_K^n)^\top}{\sqrt{d_k}}\right)f_{neg}\mathbf{W}_V^n,
  \end{aligned}
\end{equation}
where $\mathbf{W}_Q^p$, $\mathbf{W}_K^p$, $\mathbf{W}_V^p$ and $\mathbf{W}_Q^n$, $\mathbf{W}_K^n$, $\mathbf{W}_V^n$ are learnable matrices for positive and negative samples respectively, and $d_k$ is the key dimension. For contrastive learning, we group features into anchors $\{token_{pos}, f_s^v, f_c^t, f^a\}$, positives $\{f_{pos}^v, f_d^t\}$, and negatives $\{token_{neg}, f_{neg}^v, f_{bg}^t\}$. The relationships are learned through InfoNCE loss:
\begin{equation}
    \mathcal{L}_{con} = -\log\frac{\exp(f_a \cdot f_p/\tau)}{\sum_{n}\exp(f_a \cdot f_n/\tau)},
    \label{eq:con}
\end{equation}
where $f_a$, $f_p$, and $f_n$ represent anchor, positive, and negative features, and $\tau$ is the temperature. We employ modality dropout during training to prevent over-reliance on specific modalities.

\subsubsection{Dual-path Reconstruct}
To bridge the semantic-geometric gap while leveraging SAM's geometric understanding, we propose a dual-path reconstruction module that adaptively integrates semantic and geometric cues. In the semantic path, we first concatenate the global semantic tokens $(token_{pos}, token_{neg})$ and project them to obtain high-quality tokens:
\begin{equation}
    g = \sigma(\mathbf{W^T}[token_{pos}; token_{neg}] + \mathbf{b}),
\end{equation}
where $\mathbf{W} \in \mathbb{R}^{2d \times d}$ and $\mathbf{b} \in \mathbb{R}^d$ are learnable weight and bias, $\sigma$ denotes ReLU activation. The projected token $g$ serves as HQ-SAM decoder's high-quality token due to its comprehensive foreground-background knowledge. The enhanced fine-grained features $f_{pos}$ from the fusion module act as sparse semantic embeddings $emb_{sem}$ to provide local appearance details. In the geometric path, we first decompose the query image into multiple proposals $\{{P_{q,i}}\}_{i=1}^N$, and then determine the coarse location priors ($M_v$, $M_t$, $M_a$) by computing similarities with given cues::
\begin{equation}
    \begin{aligned}
    M_v &= \sigma(\frac{F_q \cdot f_s^v}{\|F_q\| \|f_s^v\|}), \\
    M_t &= \sum_{i} \mathbbm{1}[\text{sim}(F_{P_{q,i}}, f_c^t) > \delta_{t}] \cdot P_{q,i}, \\
    M_a &= \sum_{i} \mathbbm{1}[\text{sim}(F_{P_{q,i}}, f^a) > \delta_{a}] \cdot P_{q,i},
    \end{aligned}
\end{equation}
where $\delta$ is a similarity threshold, $\mathbbm{1}[\cdot]$ is the indicator function, and $\delta$ is a similarity threshold. The multi-modal priors are first encoded into geometric prompts $emb_v = \phi(M_v)$, $emb_t = \phi(M_t)$, $emb_a = \phi(M_a)$, then fused through a simple convolution block to obtain the final geometric prompt, $emb_{geo} = \text{Conv}([emb_v; emb_t; emb_a])$, where $\phi$ denotes the mask prompt encoder in SAM. The final prompt guides SAM's decoder to generate initial masks:
\begin{equation}
  \begin{aligned}
  M_{init} &= \text{SAM}_{decoder}(g, emb_{sem}, emb_{geo}, F_q), \\
  M_{pred} &= \text{Refiner}(M_{init}, F_q).
  \end{aligned}
\end{equation}

The total loss function combines segmentation objectives and contrastive learning:
\begin{equation}
    \mathcal{L}_{total} = (1-\lambda) (\mathcal{L}_{bce} + \mathcal{L}_{dice}) +\lambda \mathcal{L}_{con},
\end{equation}
where $\mathcal{L}_{bce}$ and $\mathcal{L}_{dice}$ are binary cross-entropy and Dice loss for mask prediction, and $\mathcal{L}_{con}$ is the InfoNCE contrastive loss defined in Eq.~\ref{eq:con}, with $\lambda$ set to 0.2.

\begin{table*}
  \centering
  \caption{Comparison of the proposed DFR with the current SOTA on $\text{PASCAL-}5^i$ ~\cite{shaban2017one}. Results marked in \textbf{bold} and \underline{underlined} indicate first and second-best performance respectively.}
  \label{tab:pascal_sota}
      \scalebox{0.85}{
      \begin{tabular}{cclc|cccccc|cccccc}
        \toprule[1pt]
        \multirow{2}{*}{\shortstack{\textbf{Pre-}\\ \textbf{train}}} & \multirow{2}{*}{\shortstack{\textbf{Backbone}}} & \multirow{2}{*}{\textbf{Method}} & \multirow{2}{*}{\textbf{Publication}} & \multicolumn{6}{c|}{\textbf{1-shot}} & \multicolumn{6}{c}{\textbf{5-shot}} \\ 
        & & & & $5^{0}$ & $5^{1}$ & $5^{2}$ & $5^{3}$ & \textbf{mIoU} & \textbf{FB-IoU} & $\mathbf{5^{0}}$ & $\mathbf{5^{1}}$ & $\mathbf{5^{2}}$ & $\mathbf{5^{3}}$ & \textbf{mIoU} & \textbf{FB-IoU}  \\
        \midrule[1pt] 
        \multirow{10}{*}{IN1K} & \multirow{5}{*}{RN50}  & PFENet~\cite{tian2020prior} & TPAMI'20  & 61.7 & 69.5 & 55.4 & 56.3 & 60.8 & 73.3 & 63.1 & 70.7 & 55.8 & 57.9 & 61.9 & 73.9   \\
        && ABCNet~\cite{Wang_2023_CVPR} & CVPR'23 &68.8&73.4&62.3&59.5&66.0&76.0&71.7&74.2&65.4&67.0&69.6&80.0   \\  
        &&AdaptiveFSS~\cite{wang2024adaptive} & AAAI'24&71.1&75.5&67.0&64.5&69.5&-&74.7&78.0&\textbf{75.3}&70.8&\underline{74.7}&-  \\
        &&RiFeNet~\cite{bao2024relevant} & AAAI'24&68.4&73.5&67.1&59.4&67.1&-&70.0&74.7&69.4&64.2&69.6&-  \\
        &&UMTFSS~\cite{li2024label} & AAAI'24&68.3&71.3&60.0&60.7&65.1&-&71.5&74.5&61.5&68.4&68.9&-  \\
        \cline{2-16} \\[-2.0ex]

         & \multirow{5}{*}{RN101} & PFENet~\cite{tian2020prior} & TPAMI'20  & 60.5 & 69.4 & 54.4 & 55.9 & 60.1 & 72.9 & 62.8 & 70.4 & 54.9 & 57.6 & 61.4 & 73.5   \\
        && HPA~\cite{cheng2023hpa} & TPAMI'23  & 66.4 & 72.7 & 64.1 & 59.4 & 65.6 & 76.6 & 68.0 & 74.6 & 65.9 & 67.1 & 68.9 &80.4   \\
        && ABCNet~\cite{Wang_2023_CVPR} & CVPR'23&65.3&72.9&65.0&59.3&65.6&\underline{78.5}&71.4&75.0&68.2&63.1&69.4&\underline{80.8}   \\  
        &&DCP~\cite{lang2024few} & IJCV'24&68.9&74.2&63.3&62.7&67.3&-&72.1&77.1&66.5&70.5&71.5&-  \\
        &&RiFeNet~\cite{bao2024relevant} & AAAI'24&68.9&73.8&66.2&60.3&67.3&-&70.4&74.5&68.3&63.4&69.2&- \\
        \midrule[1pt] 
        \multirow{4}{*}{SAM}
        & \multirow{4}{*}{SAM-base} & Matcher~\cite{liu2023matcher} &ICLR'24 & 67.7 & 70.7 & 66.9 & 67.0 & 68.1 & - & 71.4 & 77.5 & \underline{74.1} & \underline{72.8} & 74.0 & - \\
        &&  VRP-SAM~\cite{sun2024vrp} &CVPR'24 & 73.9 & \underline{78.3} & \underline{70.6} & 65.0 & 71.9 &-& \underline{76.3} & 76.8 & 69.5 & 63.1 & 71.4 &  -   \\
        &&   FCP~\cite{fcp}&  Arxiv'25 & \underline{74.9} & 77.4 & \textbf{71.8} & \underline{69.8} & \underline{73.2} & - & \textbf{77.2} & \underline{78.8} & 72.2 & 67.7 & 74.0 & -  \\
        &&  \cellcolor [HTML]{EFEFEF} DFR~(ours)& \cellcolor [HTML]{EFEFEF} - &\cellcolor [HTML]{EFEFEF} \textbf{76.7} &\cellcolor [HTML]{EFEFEF} \textbf{82.3} &\cellcolor [HTML]{EFEFEF} 68.0 &\cellcolor [HTML]{EFEFEF} \textbf{74.5} &\cellcolor [HTML]{EFEFEF} \textbf{75.4} &\cellcolor [HTML]{EFEFEF} \textbf{84.5} &\cellcolor [HTML]{EFEFEF} \textbf{77.2} &\cellcolor [HTML]{EFEFEF} \textbf{83.1} & \cellcolor [HTML]{EFEFEF}68.5 &\cellcolor [HTML]{EFEFEF} \textbf{76.1} &\cellcolor [HTML]{EFEFEF} \textbf{76.2} &\cellcolor [HTML]{EFEFEF} \textbf{85.2}  \\
        \bottomrule[1pt]

      \end{tabular}
      }
  \hfill

\end{table*}

\section{Experiments}
\label{sec:experiments}

\subsection{Datasets and Evaluation Metrics}
\label{ssec:datasets}

\textbf{Datasets:} The proposed method was evaluated on two distinct settings: synthetic audio-enhanced FSS and real audio-visual segmentation.

\textit{Synthetic Audio FSS:} The widely-used PASCAL-$5^i$ benchmark~\cite{shaban2017one} was utilized, which was constructed from PASCAL VOC 2012~\cite{everingham2010pascal} and augmented by the SBD~\cite{hariharan2011semantic} dataset. Following the standard protocol~\cite{tian2020prior, wang2019panet}, the 20 object categories were evenly divided into 4 folds, with 5 classes per fold. For the text modality, DeepSeek-v3~\cite{liu2024deepseek} was employed to generate fine-grained descriptive attributes for each category using carefully designed prompts that elicit discriminative visual, functional, and contextual characteristics. For the audio modality, category-specific sound effects were synthesized using AudioLDM~\cite{audioLDM}.

\textit{Real Audio-Visual Segmentation:} Further evaluation was conducted on AVS-V3~\cite{wang2024prompting}, a challenging real audio-visual segmentation dataset built upon AVSBench~\cite{zhou2022audio, zhou2024audio}. The dataset encompasses both single-source and multi-source subsets across 23 sound categories, ranging from human activities and animal sounds to vehicles and musical instruments, with comprehensive pixel-level annotations. AVS-V3 implements a rigorous evaluation framework with zero-shot and few-shot paradigms using unseen audio categories and limited training samples (1, 3, and 5 samples).

\textbf{Evaluation Protocol:} For PASCAL-$5^i$, cross-validation was employed by training on three folds and testing on the remaining fold to evaluate generalization to novel classes. For AVS-V3, the standard few-shot evaluation protocol was followed with limited support samples (1, 3, and 5 shots) and zero-shot settings for unseen categories. For both datasets, standard metrics were adopted: mean Intersection-over-Union ($\text{mIoU} = \frac{1}{C} \sum_{c=1}^{C} \text{IoU}c$) for class-wise accuracy, and Foreground-Background IoU ($\text{FB-IoU} = \frac{1}{2} (\text{IoU}\text{F} + \text{IoU}_\text{B})$) for binary segmentation quality.

\subsection{Implementation Details}
\label{ssec:implementation}

The framework was implemented in PyTorch and trained on four NVIDIA RTX 3090 GPUs. Adhering to SAM's design principle, all images were processed at a resolution of 1024×1024 using the SAM-base model with frozen parameters. For multi-modal feature extraction, LanguageBind~\cite{lbz2024} was employed to obtain unified representations for both textual descriptions and audio signals. The training process utilized the Adam optimizer with a learning rate of $1 \times 10^{-4}$, a batch size of 4 per GPU, and was conducted for 10 epochs.

\subsection{Comparison with State-of-the-Art Methods}
\textbf{Synthetic Audio-Enhanced FSS Results:} As shown in Table~\ref{tab:pascal_sota}, DFR was compared with recent methods across two backbone categories: ImageNet-pretrained (IN1K) and SAM-pretrained models. Consistent improvements were observed, with DFR achieving 75.4\% and 76.2\% mIoU in 1-shot and 5-shot settings, respectively. Notably, when using the SAM backbone, DFR outperformed recent methods such as Matcher~\cite{liu2023matcher}, VRP-SAM~\cite{sun2024vrp}, and FCP~\cite{fcp} by 7.3\%, 3.5\%, and 2.2\% in the 1-shot setting. These results highlight the effectiveness of the multi-modal framework in capturing fine-grained cross-modal correlations and improving few-shot segmentation performance.

\textbf{Real Audio-Visual Segmentation Results:} To validate the generalization capability of DFR in real audio scenarios, evaluations were conducted on the AVS-V3 dataset, as shown in Table~\ref{tab:dfr_avs}. DFR demonstrated substantial improvements across all settings, particularly achieving 59.5\% mIoU in the 0-shot scenario and 66.2\% mIoU in the 1-shot scenario, with gains of 4.8\% and 3.3\% over the previous best method, GAVS, respectively. The consistent performance improvements in both synthetic and real audio settings highlight the robustness and practical applicability of the proposed approach.

\begin{table}[htbp]
  \centering
  \caption{Comparison of the proposed DFR with the current SOTA on AVS-V3~\cite{wang2024prompting}. Results marked in \textbf{bold} and \underline{underlined} indicate first and second-best performance respectively.}
  \label{tab:dfr_avs}
  \begin{tabular}{ccccc}
  \toprule
  \multirow{2}{*}{\textbf{Method}} & \multicolumn{4}{c}{\textbf{mIoU}} \\
  \cmidrule(lr){2-5}
  & \textbf{0-shot} & \textbf{1-shot} & \textbf{3-shot} & \textbf{5-shot} \\
  \midrule
  AVSBench~\cite{zhou2022audio} & 53.0 & 56.1 & 63.2 & 63.9 \\
  AVSegFormer~\cite{wang2024avesformer} & 54.3 & 58.3 & 64.2 & 65.2 \\
  GAVS~\cite{wang2024prompting} & \underline{54.7} & \underline{62.9} & \underline{66.3} & \underline{67.8} \\
  DFR~(ours) & \textbf{59.5} & \textbf{66.2} & \textbf{67.4} & \textbf{68.1} \\
  \bottomrule
  \end{tabular}
\end{table}

\subsection{Ablations and Sensitivity Analysis}
\begin{figure}[htbp]
    \centering
    \includegraphics[width=\linewidth]{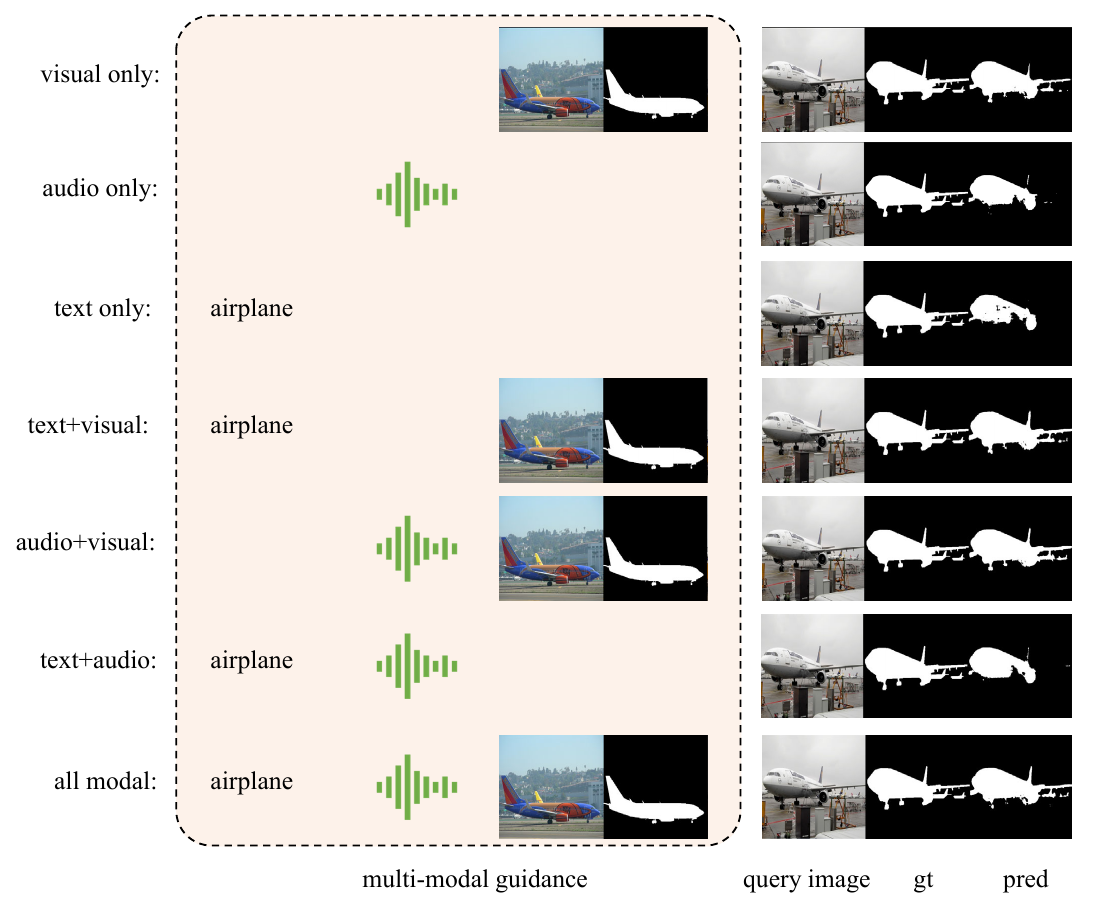}
    \caption{Visualization of few-shot segmentation on unseen classes under different modality guidance combinations: rows 1-3 show single modality guidance; rows 4-6 present dual modality guidance combinations; and the last row demonstrates all-modality guidance (combining visual, text, and audio).}
    \label{fig:modal}
\end{figure}

\label{ssec:ablation}
\textbf{Ablation on Modalities:}  Ablation studies were conducted to analyze the contribution of each modality. As shown in Table~\ref{tab:ablation_modal}, the full model incorporating all modalities achieved 76.7\% mIoU in the 1-shot setting. Among dual-modality pairs, Visual+Text exhibited the best performance (75.0\% mIoU, -1.7\%), followed by Visual+Audio (73.2\%, -3.5\%) and Text+Audio (71.5\%, -5.2\%). In single-modality tests, visual information achieved the highest performance (72.4\%, -4.3\%), outperforming text (71.3\%, -5.4\%) and audio (60.2\%, -16.5\%). These findings confirm that each modality provides complementary information, with their combination yielding optimal performance. Figure~\ref{fig:modal} visualizes segmentation results under various guidance modalities for unseen categories.

  \begin{table}[htbp]
    \centering
    \caption{Ablation study on different modality combinations. \Checkmark indicates the modality is used.}
    \label{tab:ablation_modal}
    \scalebox{1.0}{
    \begin{tabular}{cccccc}
    \toprule[1pt]
    \multirow{2}{*}{Method} & \multicolumn{3}{c}{Input Modalities} & \multicolumn{2}{c}{mIoU} \\
    \cline{2-6}
    & Visual & Text & Audio & 1-shot & $\triangledown$ \\
    \midrule[1pt]
    Full Model & \Checkmark & \Checkmark & \Checkmark & 76.7 & - \\
    Visual+Text & \Checkmark & \Checkmark &  & 75.0 & -1.7 \\
    Visual+Audio & \Checkmark &  & \Checkmark & 73.2 & -3.5 \\
    Text+Audio &  & \Checkmark & \Checkmark & 71.5 &-5.2 \\
    Visual only & \Checkmark &  &  & 72.4 & -4.3  \\
    Text only &  & \Checkmark &  & 71.3 & -5.4 \\
    Audio only &  &  & \Checkmark & 60.2 & -16.5 \\
    \bottomrule[1pt]
    \end{tabular}}
\end{table}

\textbf{Ablation on Dual-path Reconstruction:} The contributions of semantic and geometric embeddings are presented in Table~\ref{tab:ablation_feature}. The complete model achieved 76.7\% mIoU, whereas using only semantic embeddings reduced performance to 73.2\%, and using only geometric embeddings resulted in 75.1\%. These results confirm the complementary nature of semantic and geometric embeddings within the framework.

\begin{table}[htbp]
\centering
\caption{Ablation study on dual-path reconstruction components.}
\label{tab:ablation_feature}
\begin{tabular}{ccccc}
\toprule
\multirow{2}{*}{\textbf{Method}} & \multicolumn{2}{c}{\textbf{Input Features}} & \multicolumn{2}{c}{\textbf{Performance}} \\
& \textbf{Semantic} & \textbf{Geometric} & \textbf{mIoU} & \textbf{FB-IoU} \\
\midrule
Full Model & \Checkmark & \Checkmark & \textbf{76.7} & \textbf{87.5} \\
Semantic only & \Checkmark & & 73.2 & 85.0 \\
Geometric only & & \Checkmark & \underline{75.1} & \underline{86.7} \\
\bottomrule
\end{tabular}
\end{table}

\section{Conclusion}
\label{sec:conclusion}

This paper presents DFR, a novel framework that addresses the limitations of single or dual-modal approaches in few-shot segmentation by systematically integrating visual, textual, and audio modalities. The framework achieves this through three key contributions: hierarchical semantic decomposition for modality-specific feature extraction, contrastive fusion for robust cross-modal correlation learning, and dual-path reconstruction that combines semantic and geometric cues. Extensive experiments demonstrate substantial improvements over state-of-the-art methods, validating the effectiveness of tri-modal guidance for enhanced semantic understanding. Future research directions will explore additional modalities.

\bibliographystyle{IEEEtran}
\bibliography{IEEEexample}

\end{document}